\documentclass[journal]{IEEEtran}
\usepackage{cite}

\usepackage{graphicx}
\ifCLASSINFOpdf
\else
\fi
\ifCLASSOPTIONcompsoc
  \usepackage[caption=false,font=normalsize,labelfont=sf,textfont=sf]{subfig}
\else
  \usepackage[caption=false,font=footnotesize]{subfig}
\fi
\begin{document}
%
\title{A Novel Neural Network Structure Constructed according to Logical Relations}
%
%
%

\author{Gang~Wang 
\thanks{Wang is with the School of Computer Science and Technology, Taiyuan University of Technology, Taiyuan, China
(e-mail: f\_lag@buaa.edu.cn.)}
}

%
%

\markboth{}
{Wang \MakeLowercase{\textit{et al.}}: A Novel Neural Network Structure Constructed according to Logical Relations}
%



\maketitle

\begin{abstract}
To solve more complex things, computer systems becomes more and more complex. It becomes harder to be handled manually for various conditions and unknown new conditions in advance. This situation urgently requires the development of computer technology of automatic judgement and decision according to various conditions. Current ANN (Artificial Neural Network) models are good at perceptual intelligence while they are not good at cognitive intelligence such as logical representation, making them not deal with the above situation well. Therefore, researchers have tried to design novel models so as to represent and store logical relations into the neural network structures, the type of which is called KBNN (Knowledge-Based Neural Network). In this type models, the neurons and links are designed specific for logical relation representation, and the neural network structures are constructed according to logical relations, allowing us to construct automatically the rule libraries of expert systems. In this paper, the further improvement is made based on KBNN by redesigning the neurons and links. This improvement can make neurons solely for representing things while making links solely for representing logical relations between things, and thus no extra logical neurons are needed. Moreover, the related construction and adjustment methods of the neural network structure are also designed based on the redesigned neurons and links, making the neural network structure dynamically constructed and adjusted according to the logical relations. The probabilistic mechanism for the weight adjustment can make the neural network model further represent logical relations in the uncertainty.
\end{abstract}

\begin{IEEEkeywords}
Brain-inspired computing, logical representation, neural network structure, inhibitory link, probability.
\end{IEEEkeywords}

%
\IEEEpeerreviewmaketitle

\section{Introduction}
%
%
%
%
\IEEEPARstart{W}{ith} computer technology widely used for a few decades, it greatly increases the informatization and automation in every aspect of human activities, and brings the big convenience for human. As computer technology is further applied, the matters to be solved by computers become more and more complex such as advanced medical diagnosis or driverless vehicles. For these complex matters, there are a considerable amount of various conditions contained, and new conditions may appear at any time. For example, in the medical diagnosis, various symptoms may appear on a patient, such as various physical or mental features, and new symptoms may appear later for chronic or worsening diseases. People eagerly expect the coming of advanced diagnosis computer systems matching for doctors.

To solve the complex matters in medical and other fields, more and more complex computer systems need building in order to process respectively considerable conditions in the complex matters. It becomes increasingly difficult and takes considerable efforts to design manually for various conditions and for unknown new conditions in advance. This situation brings forth an urgent requirement that computers can automatically judge and make decision according to various conditions like the way the human deal with these matters. Through copying intelligent abilities of humans into computers, it can reduce humans¡¯ efforts in building the systems.

Researches on the brain as the organ of intellectual activity have been carried out like European HBP (Human Brain Project) \cite{Markram, HBP}, American BRAIN (Brain Research through Advancing Innovative Neurotechnologies) \cite{Bargmann} and Chinese Brain project \cite{Poo}. These projects are seen as another major international research projects after HGP (human genome project). Researches on brain-inspired intelligence are included as the parts of various brain projects in order to improve artificial intelligence of computers by mimicking biological brain. Multiple international IT companies like Google, IBM and Baidu have also started their own projects of brain-inspired intelligence.

\begin{figure*}[!ht]
\begin{center}
\includegraphics[width=0.9\textwidth]{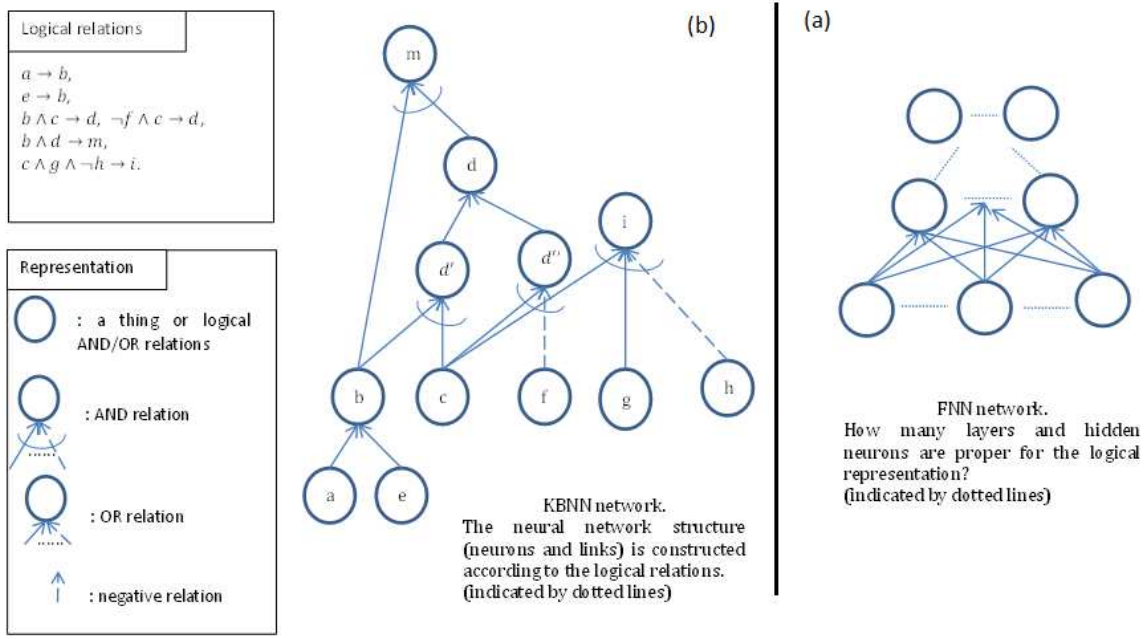}
\caption{Two different neural network models to represent logical relations}
\label{models}
\end{center}
\end{figure*}

ANN (Artificial Neural Network) has been applied successfully in perceptual intelligence like image recognition \cite{Sun, Parkhi, He} and speech recognition  \cite{Mohamed, Gehring}. These neural networks are essentially function approximators, which make them not good at cognitive intelligence such as logical representation and reasoning \cite{Garcez1, LeCun, Francois, Garnelo}. For example, to represent a set of logical relations $\{a\rightarrow b, e\rightarrow b, b\bigwedge c\rightarrow d, \neg f\bigwedge c\rightarrow d, b\bigwedge d\rightarrow m, c\bigwedge g\bigwedge \neg h\rightarrow i\}$, as shown in Figure~\ref{models}a, how is the structure of a typical FNN (Forward Neural Network) designed to represent these logical relations which means how many neurons, especially neurons in the hidden layers are needed and how many layers (indicated by dotted lines) are needed so as to represent these logical relations properly? The word "properly" means there are no extra neurons and layers with the least number in the case that the FNN can represent the above logical relations. In the process of design the structure of the FNN so as to have proper neurons and layers, the designer have to try to add neurons and layers gradually in the combination of the number of neurons and the number of layers. Obviously, it is difficult for the design of the neural network structures for logical representation, and the difficulty will increase with more and more logical relations contained. In this paper, these neural networks are called numeric neural networks. The reason for the disadvantage and difficulty is that numeric neural networks, used as function approximators, are not suitable for logical representation, and furthermore that there are no direct mapping ways from logical relations into the structure of numeric neural networks.  Although there have been some works trying to use numeric neural networks to address logical issues such as automated theorem proving \cite{Geoffrey, Cai}, the comparatively fixed predefined network structures can't make the neural network structure dynamically constructed and adjusted according to logical relations, which is an useful feature for the automatic construction of logical relations in the rule libraries of expert systems.

Logical representation and reasoning are the important cognitive intelligence since they are the base to make judgement according to various specific conditions in the form of \emph{if-then} rules, which are solved by symbolic logic, another AI branch as cognitive intelligence \cite{Luger, Michael}. In order to make ANN deal with logical issues and near to the logical ability of biological brain further, researchers have tried to study how to construct the structure of the neural network using its only two types of components of neurons and links according to logical relations, i.e. to study a novel model of ANN which can map logical relations into its structure to fulfil the purpose of representing and storing logical relation into its network \cite{Besold, Besold1}. Researchers want to design a logical kind of neural network by the way of combining symbolic logic with ANN, the two AI branches yet developing individually, and researchers expect ANN can be applied into more areas besides the areas of perceptual intelligence such as rules-based expert systems, knowledge representation and reasoning, cognitive modelling in robotics. For this purpose, there have been several approaches and applications proposed \cite{Besold, Mandziuk, Towell, Valiant, Garcez, Penning, Bowman, Gallant, Wang1}, and a neural-symbolic computation association is established. Pioneering works of designing logical neural network models are presented in \cite{Gallant, Towell, Garcez, Penning}. These models, called a type of KBNN (Knowledge-Based Neural Network) or Connectionist Expert Systems, define AND/OR neurons and positive and negative links so as to represent the basic logical connectives in logical relations of the proposition logic type. Then these models can represent complex logical relations by the composition of these neurons and links. They build AND/OR trees and form a network according to every logical relations in a rule library of an expert system, and fulfil the mapping directly from logical relations to the neural network structure. The type of KBNN represents and stores the logical relations into the connection pattern of the neural network structure.

We still use the above set of logical relations to explain further. For example, to represent the above set of logical relations, as shown in Figure~\ref{models}b, first, for the logical relation $a\rightarrow b$, KBNN creates two neurons representing the things a and b, then creates a link $l_{a, b}$ from a to b, thus representing $a\rightarrow b$. So does it for $e\rightarrow b$. At this moment, the neuron b also represent the OR relation besides representing the thing. For the logical relation $b\bigwedge d\rightarrow m$, the model creates two neurons to represent the things d and m. Then two links $l_{b, m}$ and $l_{d, m}$ are created from b to m and from d to m respectively. There are the same connection structure in the neural network between $a\bigvee e\rightarrow b$ and $b\bigwedge d\rightarrow m$. To distinguish them, the AND and OR neurons are introduced into KBNN. Here the neuron m not only represents the thing m, but also represents the AND relation. However, one AND neuron just represents one AND relations rather than multiple AND relations. Therefore, for the logical relations$\{b\bigwedge c\rightarrow d, \neg f\bigwedge c\rightarrow d\}$, intermediate or hidden AND neurons such as $d^{'}$ and $d^{"}$ are introduced to represent multiple AND relations respectively. The negative link is introduced to indicate making the inference when the thing is in the negative state. The final neural network structure of KBNN to represent the above logical relations is shown in Figure~\ref{models}b. From the structure and the constructing process, we can see the logical neural network KBNN can represent logical relations directly and easily in the contrast to the numeric neural network, which is beneficial from the components specified for representing basic logical relations. Through the composition of these neurons and links, it provides a mapping way directly from logical relations into the neural network structure. From Figure~\ref{models}b, we can see the logical relations can be reflected by the connection pattern of KBNN's network. The logical neural network KBNN fulfills the representation and storage of logical relations in the form of the neural network structure.

\begin{figure}[!t]
\subfloat[]{\includegraphics[width=0.8\columnwidth]{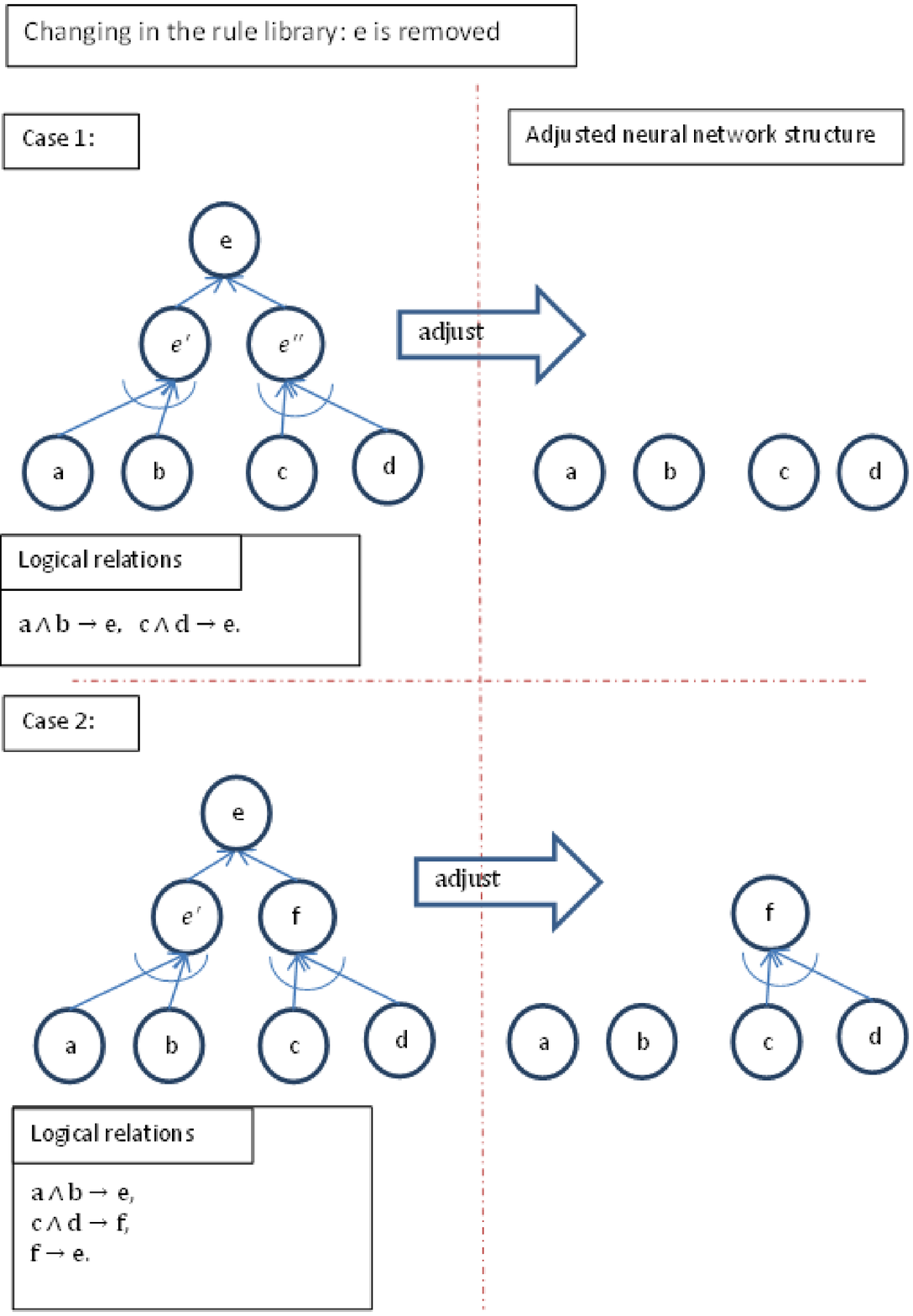}%
\label{KBNNa}}
\hfil
\subfloat[]{\includegraphics[width=1\columnwidth]{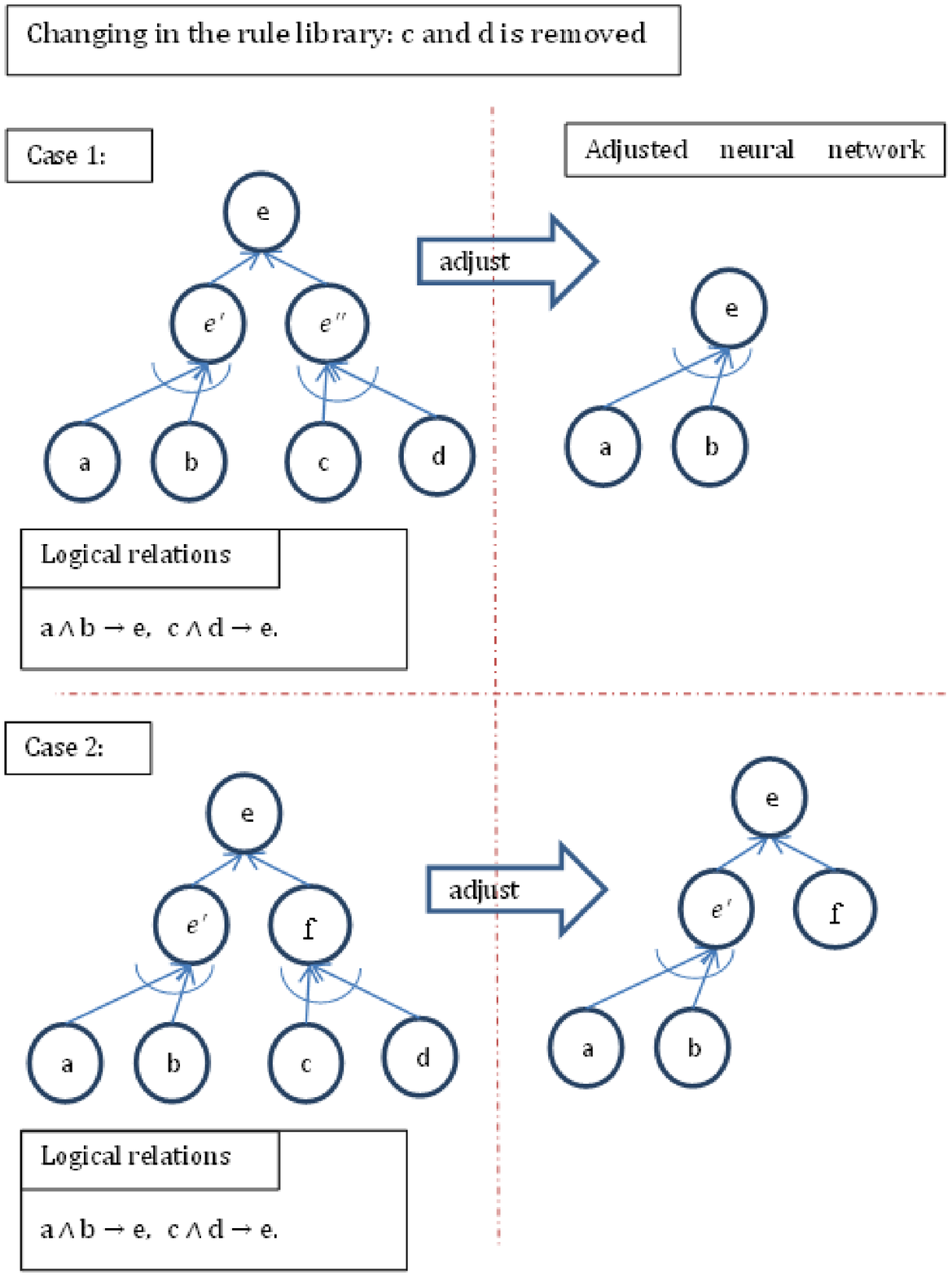}%
\label{KBNNb}}
\caption{There are multiple different adjustments of neural network structure for the same changes in the rule library, in need of considering what the neurons represent: logical relations, things or both.}
\label{KBNN}
\end{figure}

Although the KBNN type of logical neural networks can represent and store logical relations well, there is still room for improvement. Changes of logical relations may happen due to the update of the rule libraries. For example, assume the thing e is removed from the rule library as shown in Figure~\ref{KBNNa}, in Case 1, KBNN adjusts its network structure, it removes the neuron e first, and also remove the AND logical neurons $e^{'}$ and $e^{"}$ because the thing e has not existed in the rule library. By contrast, in Case 2, it removes the neuron e first, and just removes the AND logical neuron $e^{'}$. The AND logical neuron f is reserved because the neuron f is an entity neuron besides a logical neuron. For the same change in the rule library, however, KBNN is required to make multiple different adjustments in need of considering what the neurons represent: logical relations, things or both. Another example is that the things c and d are removed from the rule library as shown in Figure~\ref{KBNNb}. There are also multiple different adjustments in KBNN for the same change in the rule library. Here are just two illustrative simple examples containing a few changes of logical relation. In practice, more changes happen for logical relations in the rule libraries, making the adjustments of the neural network more difficult and troublesome. The one reason is that the neurons not only represent the things, but also represent the logical relations. It makes the designers always consider whether the neurons representing things or logical relations or both in the operations on neurons while adjusting the neural network structure.

Actually, when designing the model of the neural network, which has only two type of basic components: neuron and link, it is more natural in the way of using neurons to represent things, and using links between neurons to represent logical relations between the things represented by neurons. The style can make neurons only responsible for the representation of things while links only responsible for the representation for logical relations between things. No intercrossing on the responsibilities between neuron and links, thus avoiding the above mentioned adjustment troubles. Then it brings the advantage that: when referring to the change of logical relations between things, we just need to operate on links other than links and extra logical neurons, which reduces the difficult and trouble on adjusting the neural network structure.

Therefore, aiming for this, this paper makes further improvement based on KBNN by redesigning the neurons and links. This improvement can make links solely for representing logical relations between things and neurons solely for representing things, and then no extra logical neurons are needed. Moreover, the related construction and adjustment methods of the neural network structure are also designed based on the redesigned neurons and links. With the methods, the proposed neural network structure is dynamically constructed and adjusted according to logical relations of the proposition logic type. The probabilistic mechanism for the weight adjustment is designed to make the neural network model address the uncertainty of logical relations. Based on the three features of neural network, a neural network model called PLDNN (Probabilistic Logical Dynamical Neural Network) is proposed. It not only uses the weights of links to store information, but also uses the connection structure constructed according to logical relations. In the following, the model will be stated in details from the three features respectively.

\section{Components specified for logical relation representation}

\begin{figure}[!ht]
\begin{center}
\includegraphics[width=0.9\columnwidth]{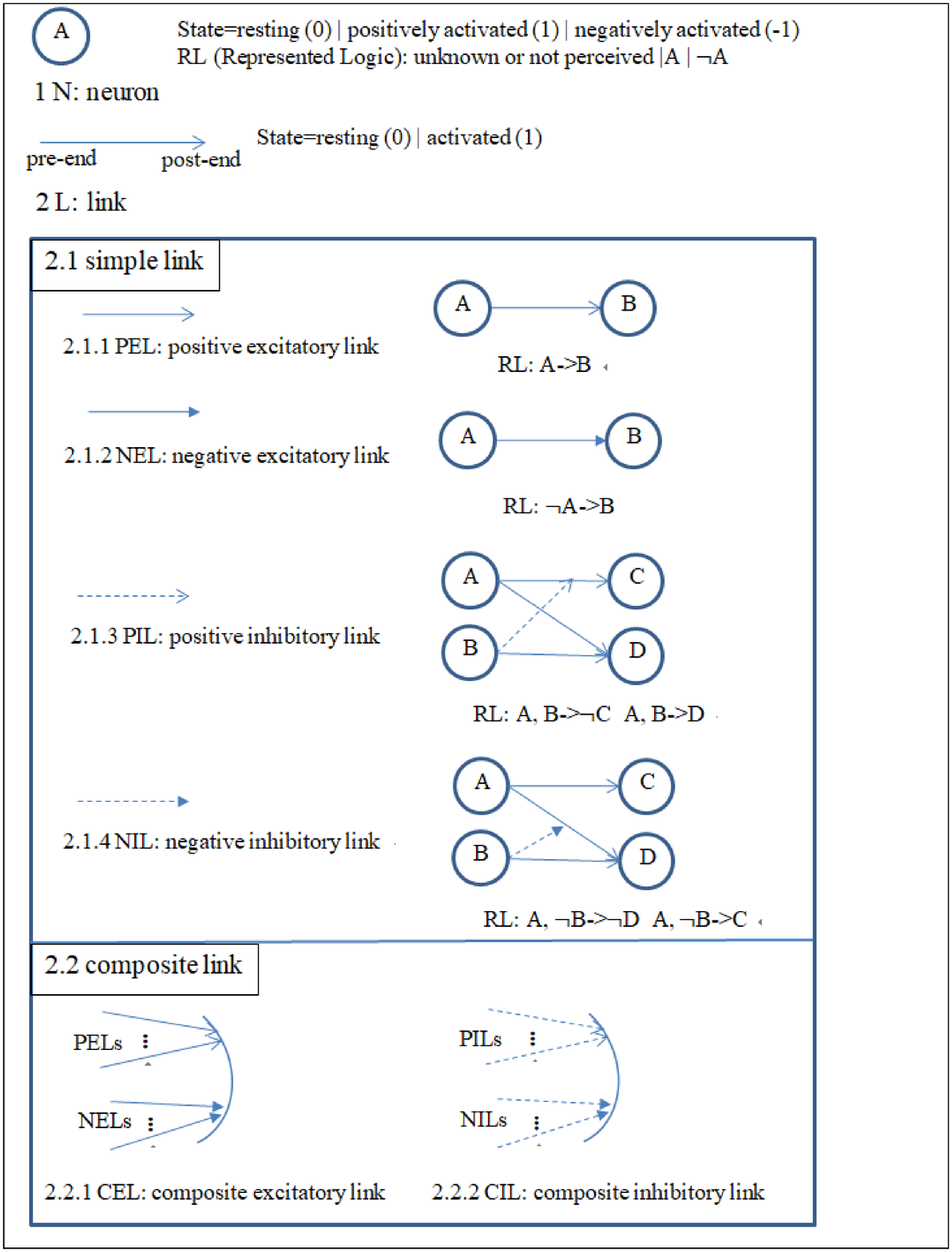}
\caption{Components of PLDNN to represent logical relations}
\label{components}
\end{center}
\end{figure}

The neurons and links in PLDNN are defined shown in Figure~\ref{components} so as to separate the neurons from the responsibilities of both the representation of things and the representation for logical relations between things, making neurons only responsible for the representation of things while links only responsible for the representation for logical relations between things To represent logical relations, there are multiple kinds of links designed in PLDNN, including excitatory and inhibitory links specified for the logical relations representation between things. Unlike the link in current numeric ANN, the pre-end of IL (Inhibitory link) connects the neuron, and its post-end connects EL (excitatory link). This connection style of IL can inhibit EL connected by it from exciting EL's post-end neuron so as to make PLDNN represent the logical relations correctly. Therefore, PLDNN has the instinctive advantages on representing logical relations using its specified components in the design aspect.

The neuron $\sigma$ is used for the thing representation and is defined by users. There are three states in $\sigma$ which are resting, positively activated and negatively activated individually, indicated by 0, 1 and -1. In general, $\sigma$ is in the resting state, borrowed from the term in the biological neuron network. When a thing A happens and perceived by PLDNN, the neuron representing A becomes positively activated to represent the logic A. Likewise, when a thing A doesn't happen and perceived by PLDNN, the neuron representing A becomes negatively activated to represent the logic $\neg A$.

Different from only numeric links contained without states in numeric ANNs, there are two types of links (ELs and ILs) designed in PLDNN so as to represent the logical relations, and the link also has states. There are two states in the link which are resting and activated, indicated by 0 and 1. When the link is in the activated state, it can make effects on its post-end. In order to represent logical relations, the two links are triggered into the activated state as follows:
\begin{itemize}
\item PEL.state =1 when its pre-end neuron.state= 1
\item NEL.state =1 when its pre-end neuron.state= -1
\item PIL.state =1 when its pre-end neuron.state= 1
\item NIL.state =1 when its pre-end neuron.state= -1
\end{itemize}

Multiple simple ELs can be put together as a composite EL to fulfill complex excitement. In the similar way, multiple simple ILs can be be put together as a composite IL to fulfill complex inhibition. Then the two composite links interact so as to represent complex logical relations. The composite links are triggered into the activated state as follows:
\begin{itemize}
\item CEL.state =1 when the states of all simple ELs contained in this CEL are 1
\item CIL.state =1 when the states of all simple ILs contained in this CIL are 1
\end{itemize}

\begin{figure}[!ht]
\begin{center}
\includegraphics[width=0.9\columnwidth]{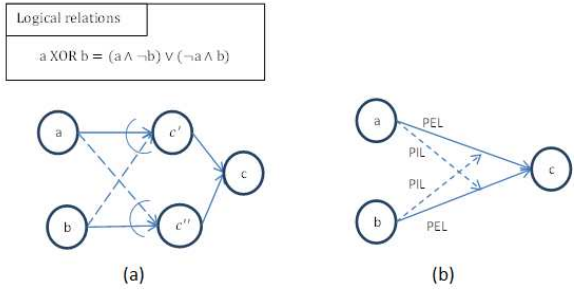}
\caption{Representing XOR logic relation}
\label{XOR}
\end{center}
\end{figure}

\emph{Representing XOR logic relation:} For the KBNN type of logical neural networks, because the neurons not only represent the things, but also represent the logical relations, the networks have to add an extra intermediate neurons to perform logic and form an intermediate layer to perform XOR operation shown in Figure~\ref{XOR}a. By contrary, by the way of making the neurons just represent things while links just represent logical relations between things, there is no intercrossing on the representing responsibilities between neuron and links, thus no extra intermediate neurons and layers are added into the neural network structure of PLDNN shown in Figure~\ref{XOR}b.

The process of performing XOR operation in the neural network structure of PLDNN is the following, considering all the four cases of the two things A and B.
\begin{itemize}
  \item If the things A and B don't happen, the neurons representing A and B are negatively activated. The link $PEL_{A, C}$ of A is not activated according to the activating condition of PEL. So does the $PEL_{B, C}$. Consequently, the neuron representing C, the post-end of $PEL_{A, C}$ and $PEL_{B, C}$, is not activated indicating that computation result of A XOR B is FALSE, i.e. the thing C is reasoned not to happen afterwards.
  \item If the things A doesn't happen and B happens, the neuron representing A is negatively activated and the neuron representing B is positively activated. The link $PEL_{A, C}$ of A is not activated. $PEL_{B, C}$ is activated. Consequently, the neuron representing C is activated by $PEL_{B, C}$, indicating that computation result of A XOR B is TRUE, i.e. the thing C is reasoned to happen afterwards.
  \item If the things A happens and B doesn't happen, the neuron representing A is activated and the neuron representing B is positively activated. The computation is similar to the case that things A doesn't happen and B happens, just exchanging their positions. In this case, the neuron representing C is activated by $PEL_{A, C}$, indicating that computation result of A XOR B is TRUE, i.e. the thing C is reasoned to happen afterwards.
  \item If the things A and B happen, the neurons representing A and B are positively activated. The link $PEL_{A, C}$ of A is activated. So does the $PEL_{B, C}$. The link $PIL_{A, PEL_{B, C}}$ of A is activated according to the activating condition of PIL. The inhibiting effect of $PIL_{A, PEL_{B, C}}$ works in the activated state, inhibiting $PEL_{B, C}$ from activating C. So does the $PIL_{B, PEL_{A, C}}$. Eventually, the neuron representing C is not activated indicating that computation result of A XOR B is FALSE, i.e. the thing C is reasoned not to happen afterwards.
\end{itemize}
\section{Automatic Construction of Neural Network for Logical Relations along with Events}
In reality, events continuously happen all the time, which contains a kind of sequential relation or causal relation between one event $e_{i}$ and the next event $e_{i+1}$, represented in the form of the logical relation $e_{i}\rightarrow e_{i+1}$. More specifically, $\{t_{i,1}, t_{i, 2}, ..., t_{i, m}\}\rightarrow \{t_{i+1, 1}, t_{i+1, 2}, ..., t_{i+1, n}\}$, where the m things $t_{i,1}, t_{i, 2}, ..., t_{i, m}$ happen at the same time in $e_{i}$ while the n things $t_{i+1, 1}, t_{i+1, 2}, ..., t_{i+1, n}$ happen at the same time in $e_{i+1}$. For example, the industrial process flow of producing sulfuric acid $H_{2}SO_{4}$ from pyrite $FeS_{2}$ as the raw material is as follows: 1)Burning $FeS_{2}$, then get $SO_{2}$; represented in the above form, it is $\{FeS_{2}, O_{2}\}\rightarrow \{Fe_{2}O_{3}, SO_{2}\}$, 2)Oxidizing $SO_{2}$, then get $SO_{3}$; it is $\{SO_{2}, O_{2}\}\rightarrow \{SO_{3}\}$, 3)Absorbing $SO_{3}$, then get $H_{2}SO_{4}$; it is $\{SO_{3}, H_{2}O\}\rightarrow \{H_{2}SO_{4}\}$. In this paper, the neural network structure will be designed to be constructed synchronically following the happening of these events, so as to represent and store the sequential relations or causal relations between events. Additionally, this construction method can also make the neural network fulfill the incremental representation and storage of logical relations along with time.

In the following, the paper presents how to construct the neural network N following the happening of the events through the example of the production of $H_{2}SO_{4}$. It also presents how to solve the correct representation of the logical relations in the format of neural network when multiple logical relations are represented into a single neural network.
\begin{figure}[!ht]
\begin{center}
\includegraphics[width=0.9\columnwidth]{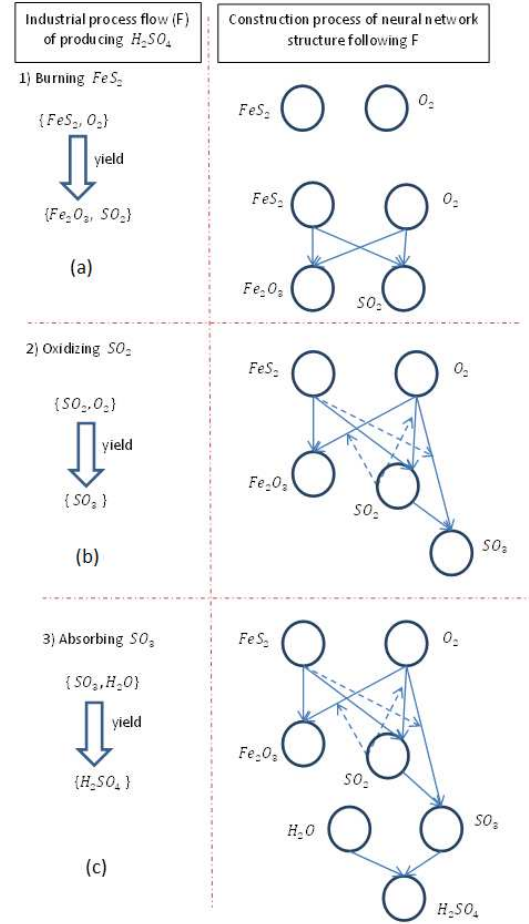}
\caption{The neural network structure is constructed according to logical relations of the production of $H_{2}SO_{4}$}
\label{constructionH2SO4}
\end{center}
\end{figure}
For the sub process of burning $FeS_{2}$, as shown in Figure~\ref{constructionH2SO4}a, two neurons are created to represent the reactant things $FeS_{2}$ and $O_{2}$ in the neural network N. When the products $Fe_{2}O_{3}$ and $SO_{2}$ appear, another two neurons are created to represent them, and the directed links are created from the reactants to the products, thus representing the logical relation between the neighboring events $\{FeS_{2}, O_{2}\}\rightarrow\{Fe_{2}O_{3}$, $SO_{2}\}$. Next, for the sub process of oxidizing $SO_{2}$, the construction process is the same way with that process in burning $FeS_{2}$. As shown in Figure~\ref{constructionH2SO4}b, First, two neurons are created for representing the reactants things $SO_{2}$ and $O_{2}$. Because the neurons represent them have existed in N, no new neurons do not need creating. In oxidizing $SO_{2}$, the new thing $SO_{3}$ appears. N creates the new neuron representing $SO_{3}$, and also creates the directed links from the reactants to the products.

Readers have noticed that the constructed neural network at present cannot represent and store the above two logical relations, or memorize them in other words. When the things $FeS_{2}$ and $O_{2}$ appear again, N cannot reason the next event rightly according to its present neural network structure. Specifically, when N will perceives the event $\{FeS_{2}, O_{2}\}$ happens again in future, the neurons $FeS_{2}$ and $O_{2}$ will be activated, and they will excite their next neurons through the directed links to make N perform the reasoning. The neuron $FeS_{2}$ will excite the neurons $Fe_{2}O_{3}$ and $SO_{2}$ by following its links, and the neuron $O_{2}$ will excite the neurons $Fe_{2}O_{3}$, $SO_{2}$ and $SO_{3}$ by following its links. After these information transmission in the network, N reason wrongly that the next event is $\{Fe_{2}O_{3}, SO_{2}, SO_{3}\}$. Actually, only the things $Fe_{2}O_{3}$ and $SO_{2}$ will appear after $FeS_{2}$ and $O_{2}$ appear. Therefore, to make N represent the right logical relations and reason rightly, it needs some mechanisms to control the directions of information flow in the neural network. Here, N should prevent the neuron $O_{2}$ from exciting the neuron $SO_{3}$ in the condition that $FeS_{2}$ is together with $O_{2}$. After the analysis, it is found that the determining factor is $FeS_{2}$, and that the controlled factor is the link from $O_{2}$ to $SO_{3}$ as a path of information transmission. Therefore, based on the analysis, a new type of link is introduced into the neural network model to cut off the information transmission over the link from $O_{2}$ to $SO_{3}$ conditionally.

Instead of connecting the neurons at both ends, which is the traditional design style of neural network models, its one end connects the link \{$O_{2}$, $SO_{3}$\}, and its other end connects the neuron $FeS_{2}$. When $FeS_{2}$ appears, this link takes effects, and stop the link \{$O_{2}$, $SO_{3}$\}, the information pathway of the neuron $O_{2}$ to excite the neuron $SO_{3}$. The neuron $O_{2}$ only excites the neuron $SO_{2}$ through the link \{$O_{2}$, $SO_{2}$\}. Then the neural network reasons rightly that $SO_{2}$ will appear after $FeS_{2}$ and $O_{2}$ appear, thus representing and store the logical relation $\{FeS_{2},  O_{2}\}\rightarrow\{SO_{2}, Fe_{2}O_{3}\}$ rightly using this connection structure. It is the same principle for the logical relation $\{SO_{2}, O_{2}\}\rightarrow\{SO_{3}\}$.

According to the effects of this link, in this paper, it is called the inhibitory link (IL) while the link, which connects neurons at both ends, is called the excitatory link (EL).

\emph{The principle behind introducing the new inhibitory link is to make the neural network can represent the logical relation $\rightarrow\neg$. Then interacting with excitatory links, the neural network consequently has the ability of knowing what don't happen and what happen in the next event based on specific conditions.} The meanings of these two sentences are explained by using the example of industrial production of $H_{2}SO_{4}$ too. When the things $FeS_{2}$ and $O_{2}$ appears, the neurons $FeS_{2}$ and $O_{2}$ are thus activated indicating the neural network N perceives the two things happen, then the two neurons will excite the directed linked neurons through their connected excitatory links. The neuron $FeS_{2}$ will excite the neurons S$O_{2}$ and $Fe_{2}O_{3}$, and the neuron $O_{2}$ will excite the neurons $SO_{2}$, $Fe_{2}O_{3}$ and $SO_{3}$. At the same time, in the effect of the IL of the neuron $FeS_{2}$, the neuron $O_{2}$ is prevented from exciting $SO_{3}$ through the $EL_{O_{2}, SO_{3}}$, thus representing the logical relation $FeS_{2}\bigwedge O_{2}\rightarrow\neg\ \ SO_{3}$. Consequently, N knows that when $O_{2}$ appears with $FeS_{2}$, $SO_{2}$ rather than $SO_{3}$ will appear next, and it represents and stores the logical relation $FeS_{2}\bigwedge O_{2}\rightarrow SO_{2}\bigwedge Fe_{2}O_{3}$ in the format of the neural network structure, i.e. the connection patterns of neurons and links.

N continues memorizing the logical relations when $H_{2}O$ appears in absorbing $SO_{3}$. As shown in Figure~\ref{constructionH2SO4}c, first, a new neuron is created to the new thing $H_{2}O$. In absorbing $SO_{3}$, $H_{2}SO_{4}$ appears, herein another neuron is created to represent $H_{2}SO_{4}$, and the directed links from the reactants to the product are created, thus representing and storing another new logical relation $SO_{3}\bigwedge H_{2}O\rightarrow H_{2}SO_{4}$ into its network structure.

\begin{figure}[!ht]
\begin{center}
\includegraphics[width=0.9\columnwidth]{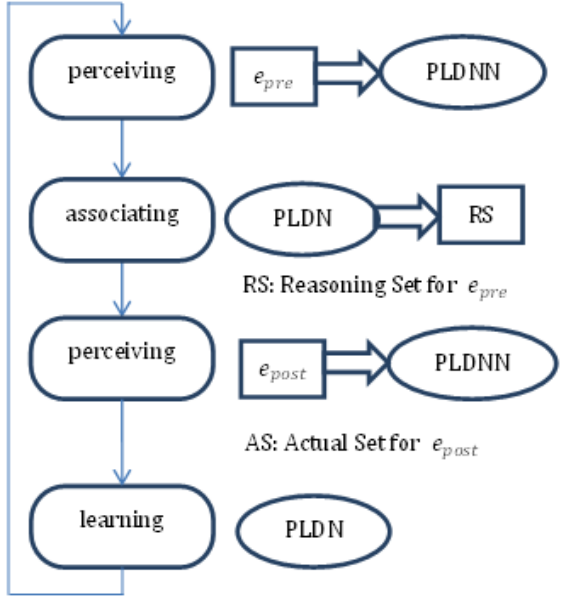}
\caption{Four stages of the working process of PLDNN}
\label{PLDNNstages}
\end{center}
\end{figure}

Illustrated through the above example, to make N constructed along with the happening of the events, the process is designed as follows:

As shown in Figure~\ref{PLDNNstages}, The process includes 4 stages: perceiving the event, associating, perceiving the next event and learning. 1) When N perceives an event $e_{pre}$, neurons representing the things in $e_{pre}$ are activated. If the neurons representing these things do not exist, new neurons are created and become activated. 2) In the associating stage, N reasons what is the next event after $e_{post}$ by the interaction the activated neurons through their ELs and ILs. The next event reasoned is a set type, named RS (Reasoning Set) in this paper. 3) N perceives the next event $e_{post}$, thus indicating the logical relation $e_{pre}\rightarrow e_{post}$. The things in the next event $e_{post}$ compose AS (Actual Set) in contrast with RS because $e_{post}$ actually happens after $e_{pre}$. 4) N adjusts its neural network structure according to the consistency between RS and AS, reducing the difference between the reasoning result and actual result to achieve the right reasoning. The adjustment includes the creation of neurons and links to change its internal connection structure so as to represent and store the logical relations into its neural network. Specifically, regarding the consistency between RS and AS, neurons can be completely divided into four cases shown in Table~\ref{consistency}. For Case 1, it means that, for a neuron not existing in RS and AS, N reasons a thing represented by the neuron does not happen, and the thing actually does not happen, that is to say N reasons rightly a thing does not happen. For Case 2, it means that, for a neuron not existing in RS and existing in AS, N reasons a thing represented by the neuron does not happen, and the thing actually happen, that is to say N reasons wrongly a thing does not happen. It is the similar meanings for the other cases. For the wrong cases, PLDNN add ELs or ILs to adjust the network structure so as to represent and store new logical relations to improve the preciseness of reasoning. For the right cases, PLDNN just keeps the present network structure. More details about the construction and adjustment algorithms implemented in Java are seen in appendix~\ref{algorithm}.

N continues performing the four stages again and again along with the happening of the events. In this loop, N continues unstoppably adjusting its network structure by creating neurons and links according to the logical relations. The dynamical characteristic of its network structure makes it adaptive for representing and storing new logical relations into the neural network and makes it also have the ability of incremental learning.
\begin{table}[!ht]
  \caption{Consistency between RS and AS}\label{consistency}
  \centering
  \begin{tabular}{|c||c||c||c|}
  \hline
  case & RS & AS & consistency\\
  \hline
  1 & 0 & 0 & yes\\
  \hline
  2 & 0 & 1 & no\\
  \hline
  3 & 1 & 0 & no\\
  \hline
  3 & 1 & 1 & yes\\
  \hline
 \end{tabular}
\end{table}
\section{Representing Uncertainty of Logical relations in Neural Network Structure}
Uncertainty generally exists due to the incomplete information caused by various limits such as the perceiving range. There is no exception for logical relations. For example, in the area of medical diagnosis, when the temperature of a person is higher than normal, it is a common symptom of multiple diseases. We cannot determine which disease the person certainly catches just based on the temperature. However, we often don't get the complete information for the judgment. In this situation, we should know what things happen with how much probability for second best.

To addressing the uncertainty of logical relations so as to increase the practicability of the neural network N, on the hand, it requires N have the adaptivity of continually updating the neural network structure to represent and store more and more clear logical relations in the process of continually perceiving the flow of events and getting more information. On the other hand, in the situation of just having incomplete information, it requires N knows what things happen next with how much probability, and knows which happen most likely and so on in the condition of uncomplete logical relations.

In the flow of the happening of the events, frequency counts of the events can be gotten. It is another piece of objective information besides the sequencing orders of the events, which can be used to measure the probability. Therefore, it is nature that N uses the frequency-counting method to measure the probability so as to have the ability of knowing what things and how likely happen next.
\begin{figure}[!ht]
\begin{center}
\includegraphics[width=0.9\columnwidth]{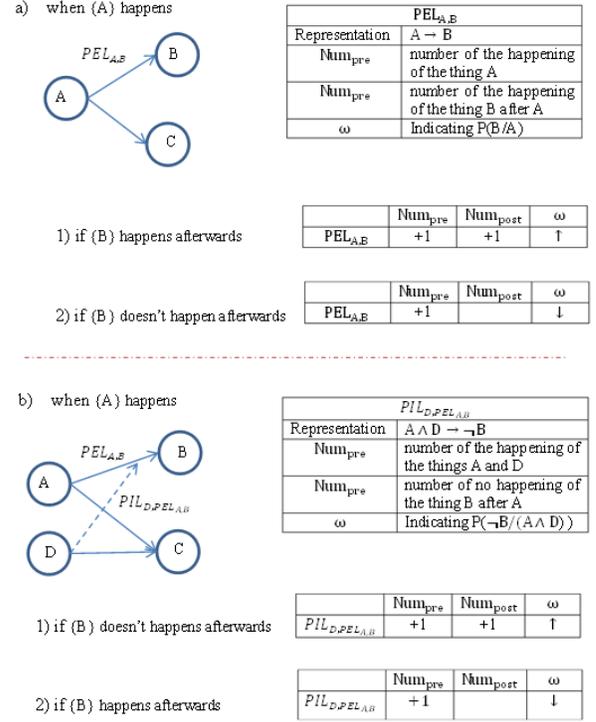}
\caption{Weight adjustment based on frequency-counting method}
\label{Weightadjust}
\end{center}
\end{figure}

In the following, the probabilistic mechanism in the proposed neural network model will be explained in Figure~\ref{Weightadjust}. When there is uncertainty in logical relations, then there must be multiple ELs in the pre-end neurons in the neural network structure. As shown in Figure~\ref{PLDNNstages}a, for the simple logical relations $\{A\rightarrow B, A\rightarrow C\}$, there are two possibly happening events {B} or {C} after the event {A} happens. When {A} happens, the neuron representing A excites its directed neurons through its two excitatory links. In this situation, N doesn't know whether the next event is {B} or {C} if the weights of the two links are none. Therefore, if {B} actually happens, the weight of $EL_{A, B}$ should be increased to guide N reasons that B happens more likely after A. Therefore, a counter $Num_{post}$ is put into $EL_{A, B}$. It will increase by 1 when the thing B represented by the post-end neuron of $EL_{A, B}$ happens. Similarly, $EL_{A, B}$ also has another counter $Num_{pre}$ to count the number of the happening of the thing A represented by the pre-end neuron of $EL_{A, B}$. Then $Num_{post}/Num_{pre}$ is used as the weight of the link. This weight will increase when {B} happens after {A} happens, and it will decrease when {B} doesn't happen after {A} happens.

In addition, as shown in Figure~\ref{PLDNNstages}b, when the event {A, D} happens, the neuron representing A excites its directed neurons through its two excitatory links. The neuron representing D excites its directed neurons through its excitatory links, and the neuron D also prevents the neuron A from exciting B through its $IL_{D, EL_{A,B}}$, which makes the inhibitory effect on $EL_{A,B}$. In the interaction between the neurons through excitatory and inhibitory links, N reasons C will happen next after A and D happen. If {C} happens next, in the learning stage, besides updating ELs¡¯ weights in N as the above mentioned ways, $IL_{D, EL_{A,B}}$ also has $Num_{post}$, and increases by 1 when B, the post-end neuron of the inhibited $EL_{A,B}$, doesn't happen. $Num_{pre}$ of $IL_{D, EL_{A,B}}$ also increase by 1 when the thing D happens. On the other hand, if B happen next, the $Num_{post}$ of $IL_{D, EL_{A,B}}$ doesn't increase because B happens. The $Num_{pre}$ of $IL_{D, EL_{A,B}}$ still increase by 1 because D happens. With $Num_{post}/Num_{pre}$ used as the weight of IL, it increases when B doesn't happens after A happens, strengthening the inhibitory effect of the inhibitory link. On the other hand, it decreases when B doesn't happen after A happens, weakening the inhibitory effect.

In the interaction between neurons through excitatory and inhibitory links whose weights continuously update along with the happening of events, N has the ability to know what things happen next with the likelihood by performing this probabilistic mechanism. The general description of this probabilistic mechanism is as following: for an EL, if the event represented by its pre-end neurons happen, then its counter $Num_{pre}=Num_{pre}+1$, and if the things represented by its post-end neurons happen, then its counter $Num_{post}=Num_{post}+1$. The ratio $Num_{post}/Num_{pre}$ is used as the weight of the EL. For an IL, if the event represented by its pre-end neurons happen, then its counter $Num_{pre}$=$Num_{pre}$+1, and if the things represented by its post-end neurons don't happen, then its counter $Num_{post}$=$Num_{post}$+1. The ratio $Num_{post}$/$Num_{pre}$ is used as the weight of the IL. The meaning behind the weight design is to let the weight of the links further represent and store the conditional probability. For an EL, its weight represents and stores the conditional probability $P(t_{post}/(t_{1}, ..., t_{i}, \neg t_{i+1}, ..., \neg t_{n}))$ on the base that the EL represents the logical relation $t_{1}\bigwedge ... t_{i}\bigwedge\neg t_{i+1}\bigwedge ... \neg t_{n}\rightarrow t_{post}$. For an IL, its weight represents and store the conditional probability $P(\neg t_{post}/(t_{1}, ..., t_{i}, \neg t_{i+1}, ..., \neg t_{n}))$ on the base that the IL represents the logical relation $t_{1}\bigwedge ..., t_{i}\bigwedge \neg t_{i+1}\bigwedge ... \neg t_{n}\rightarrow\neg t_{post}$. With the representation and storage of the probability into the neural network, the neural network has the ability of knowing what things happen next with how much probability.

The advantages of this probabilistic mechanism are 1) the weight adaption is determined by the external event flow rather than manual settings. Two counters count the numbers of the happening of the events, and the weights are updated synchronically along with the flow of the events. N memorizes the probability into its network structure, then it can reason what things happen next in the condition of uncertainty through the probability. 2) the weights of links holding the probability have proper excitatory and inhibitory effects to make N reason more near to the actual results. That means, for an EL, the increment of the weight will strengthen its exciting effect on neurons to let N reason the happen of their represented thing more likely. For an IL, the increment of the weight will strengthen its inhibitory effect on neurons to let N reason their represented thing doesn't happen more likely. The weight alteration works similar to Hebb's rule \cite{Hebb}. In this way of the weight adjustment, N will try to excite the neurons connected by the EL with the high probability, and inhibit the neurons connected by the IL with the high probability so as to make the reasoning result of the next happening event more near to the actual result. The algorithm implementation of the weight adjustment can be seen in appendix~\ref{algorithm}.

The nature of weight adjustment is to count the numbers of the things appear or not appeared involved in the links. For an $EL_{(t_{1}, ..., t_{i}, \neg t_{i+1}, ..., \neg t_{n}), t_{post}}$, the link counts the number of the situation that the things $t_{1}, ..., t_{i}$ happen and $t_{i+1}, ..., t_{n}$ don't happen, and stores it in $Num_{pre}$. At the same time, it counts the number of the happening of the thing $t_{post}$ in the above situation, and stores it in $Num_{post}$. For an $IL_{(t'_{1}, ..., t'_{i'}, \neg t'_{i'+1}, ..., \neg t'_{n'}), EL_{(t_{1}, ..., t_{i}, \neg t_{i+1}, ..., \neg t_{n}), t_{post}}}$, the link counts the number of the situation that the things $t'_{1}, ..., t'_{i'}$ happen and $t'_{i'+1}, ..., t'_{n'}$ don't happen when the things $t_{1}, ..., t_{i}$ happen and $t_{i+1}, ..., t_{n}$ don't happen, and stores it in $Num_{pre}$. At the same time, it counts the number of no happening of the thing $t_{post}$ in the above situation, and stores it in $Num_{post}$.
\section{Verification and Demonstration}
In the following, Subsection \ref{Verification} will carry out the experiments to to verify the feasibility of this neural network model in representing logical relations by using the datasets from different domains. Subsection \ref{Demonstration} will demonstrate the characteristics of the model through an intuitive demo in representing logical relations in appendix~\ref{Supplements}. The logical relations of the illustrative rule library used in the demonstration also usually appear as example in AI related materials and textbooks like Haykin's Neural Networks and Learning Machine \cite{Simon}.
\subsection{Verification of the Feasibility in Representing Logical Relations}
\label{Verification}
After the above theoretical design of the neural network model, the experiments are carried out to verify the feasibility of this neural network model in representing logical relations in a rule library by using two datasets from different domains: Zoo \cite{UCI,Forsyth} and Breast Cancer Wisconsin (Original) \cite{Mangasarian} datasets from UCI dataset library. Zoo is the dataset about the animals and their feature attributes. In the experiments, Zoo is tailed into 3 test sets in the incremental sizes, which are the set with 10 attributes and 20 kinds of animals, the set with 15 attributes and 40 kinds of animals, and the set with 20 attributes and 60 kinds of animals. Three data sets in different sizes are used to test whether the model still workable along with the data increments.

For the biggest Zoo data set, it contains 20 attributes, the attribute "leg" has six values, "animal type" has seven values, and "size" has three values. The rest 17 attributes have two values. The combination of the 20 attributes will form an enormous feature space by $2^{17}\times6\times7\times3$. Zoo will test whether the neural network model can represent and store the specific combinations of the logical relations between the animals and their corresponding feature attributes in the dataset. If the neural network model can recall the animals rightly given the feature attributes in every record in the dataset, then it proves that the model has represented and stored the logical relations of Zoo into its network structure. It will test the ability that whether the neural network structure of the model can be constructed according to the specific combinations contained by Zoo among the enormous feature space.

The experimental process is as follows: first, the datum of the animal feature attributes such as "hair", "feathers" and "eggs" in the first record $\{v1_{hair}, v1_{feathers}, ...\}$ is given to the model, then the model creates neurons representing these attributes when seeing them first. For the multi-value attributes such as "leg" of \{0, 2, 4, 5, 6, 8\} indicating the number of animals' legs, the neurons will be created according to the number of values of the attributes. For the attribute "leg", six neurons are created to representing every value. After giving the datum of the animal features, the datum of "animal\_name" in the first record $\{v1_{animal\_name}\}$ is given later as the next event to represent the logical relation $v1_{hair}\bigwedge v1_{feathers}\bigwedge...\rightarrow v1_{animal\_name}$. According to the construction process of the neural network in section{}, the model constructs neurons and links according to the first record so as to represent and store the logical relation contained in the first record into the neural network structure. This experimental process is repeated one record by one record in Zoo until the model memorizes all the logical relations of Zoo and recalls them all or stops in the limited steps if it cannot recalls them all. Through the experiments, the experiment result in Table~\ref{Results} shows that the model is workable in representing and storing logical relations. The neural network model constructs the corresponding network structure according to Zoo data set, and it can represent and store the specific combinations of the logical relations of Zoo among the enormous feature space.
\begin{table}[!h]
\caption{Results of representing and storing logical relations}
\label{Results}
\begin{center}
\begin{tabular}{|c||c||c|}
\hline
Num(attributes) & Num(animals) & recalling degree\\\hline
10 & 20 & all \\\hline
15 & 40 & all \\\hline
20 & 60 & all \\\hline
\end{tabular}
\end{center}
\end{table}

To verify the university and practicability of the neural network model, Breast Cancer Wisconsin (Original) Data Set, which is from the medical domain, is used as test data. It contain 699 real instances from the University of Wisconsin Hospitals and publicly available from UCI dataset library. The data Set records the values of multiple feature attributes of the benign and malignant instances, and all the feature attributes are the type of integer, from 1 to 10. Similar with the experiment process for Zoo data set, for each record, the values of the feature attributes are given to the neural network. According to the construction process of the neural network, the model treats this an event $e_{pre}$, and activates the neurons representing the things in $e_{pre}$. Next, the result of "benign" or "malignant" is given to the model. The model treats the result as the next event $e_{post}$ after $e_{pre}$, and then creates or adjusts the links between neurons to update the neural network structure so as to represent and store the logical relation between the feature attributes and the tumors. Through the experiment, it shows that the model has represented and stored the logical relations of Breast Cancer dataset into its network structure, which means the neural network can recall the results of "benign" or "malignant" rightly given the values of the feature attributes for every records in the dataset. The figures of the learning neural network structures are complex and take too much space to be shown.
\subsection{Intuitive Demonstration to Explain the Characteristics of PLDNN}
\label{Demonstration}
\subsubsection{Neural network structure constructed according to the logical relations}
\begin{figure}[!t]
\centering
\subfloat[]{\includegraphics[width=0.9\columnwidth]{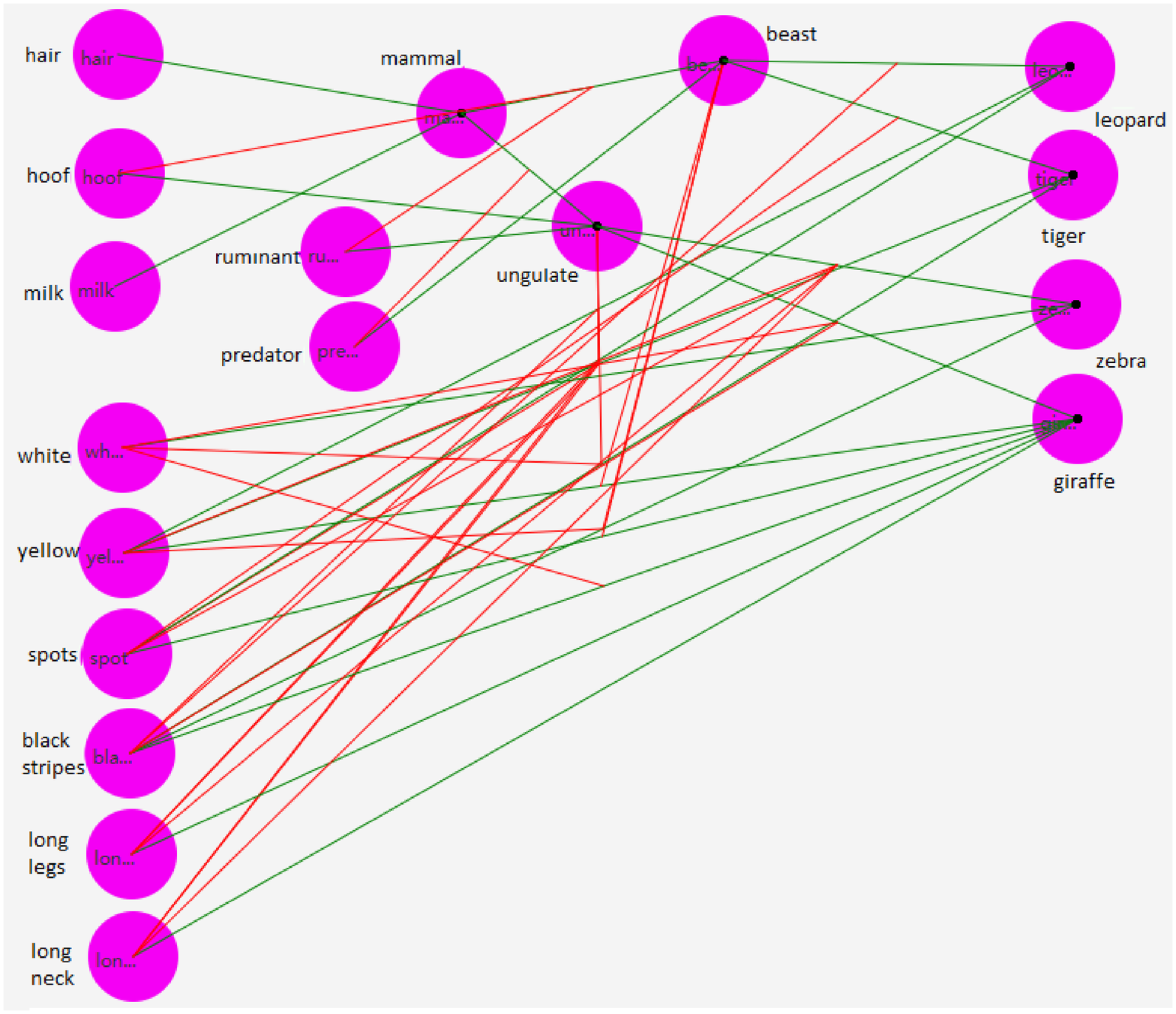}%
\label{mammal}}
\hfil
\subfloat[]{\includegraphics[width=0.9\columnwidth]{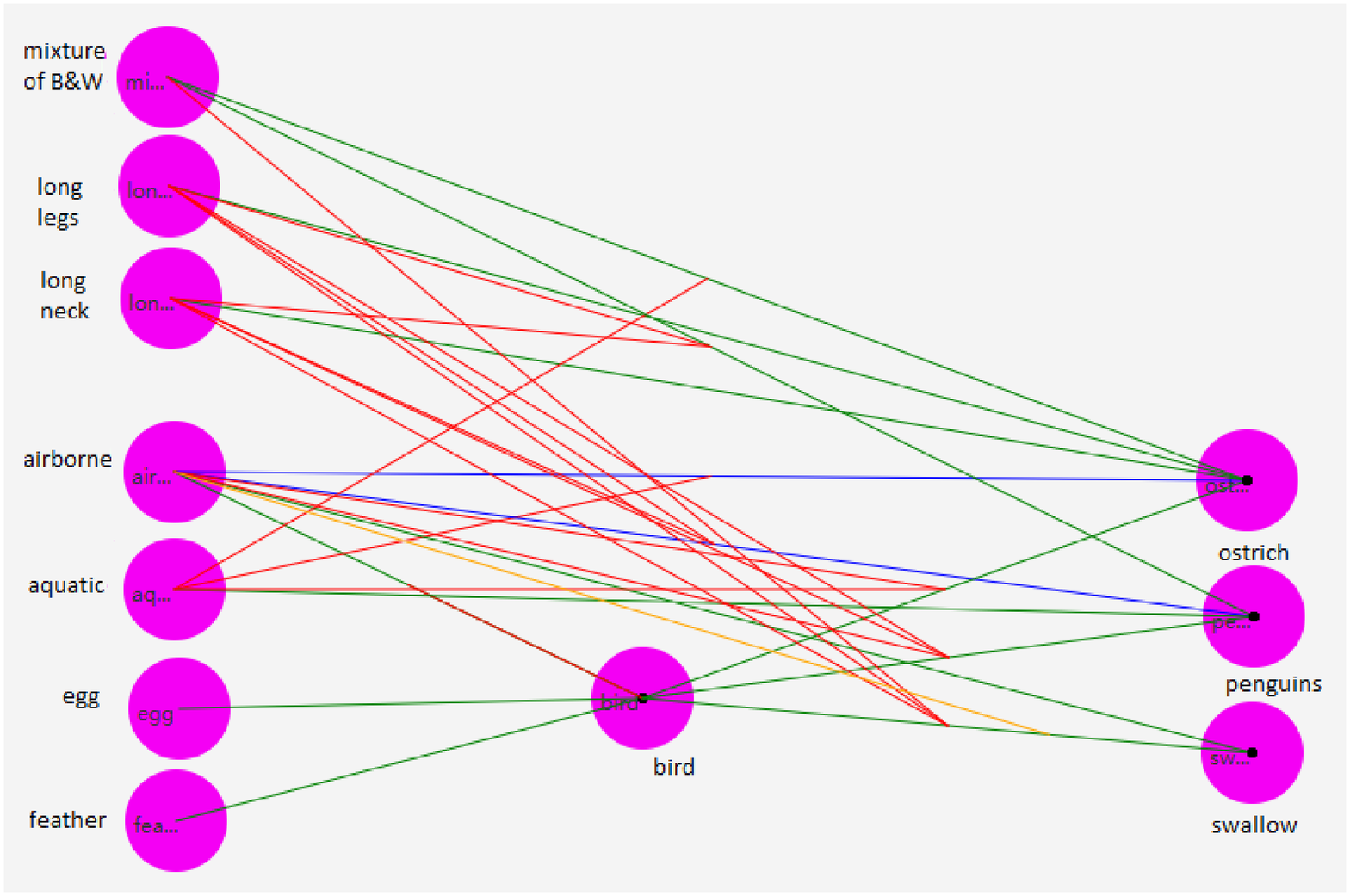}%
\label{bird}}
\caption{Neural network structures to represent logical relations of mammals or birds}
\label{animal}
\end{figure}

From Figure~\ref{mammal}, we can see the neural network structure expresses the logical relations in the illustrative rule library, including the hierarchy between the logical relations. For example, according to the illustrative rule library, the logical relation from "hair" to "mammal" forms a layer while the logical relation from "hair" to "leopard" forms three layers which are from "hair" to "mammal", from "mammal" to "beast", and from "beast" to "leopard". At the same time, as we can see intuitively, that the neural network also has the same hierarchies expressing this hierarchic relations of the logical relations. This characteristic of expressing the logical relations directly in the neural network structure comes from the construction method of the model. The neural network structure is constructed according to the logical relations. The neurons and links are created dynamically along with the happening of the events. The model uses the dynamical neural network structure and components specified for logical representation to memorize the logical relations into its neural network structure. This way is different from contemporary neural network models which have a comparatively fixed pre-defined network structure. This paper calls this characteristic of the neural network structure as growing like the logical relations. This makes the neural network structure adaptive for representing new logical relations because the network structure can change dynamically according to the logical relations. For example, if a new logical relation "If an animal is a Giraffe, then it is in Africa." is added, the model will create a neuron representing Africa, and add a excitatory link from the neuron "Giraffe" to the neuron "Africa" to represent and store this logical relation into its neural network. The rest structure will keep unchanged since the other logical relations are also unchanged. The update of the internal network structure is linked with the changes of logical relations of the external world so as to achieve the synchronization.
\subsubsection{Integration of multiple existing learned neural network}
When various neural networks are developed by different inactions, there is a requirement to integrate them into one rather than developing a new one, e.g. imaging that the scenario integrating various neural networks for recognizing different diseases of the same organ. For the proposed neural network model, based on the characteristic of the dynamical network structure adaptive for representing logical relations, it is easy and useful to integer multiple existing learned neural network into one network.

\begin{figure}[!ht]
\begin{center}
\includegraphics[width=0.9\columnwidth]{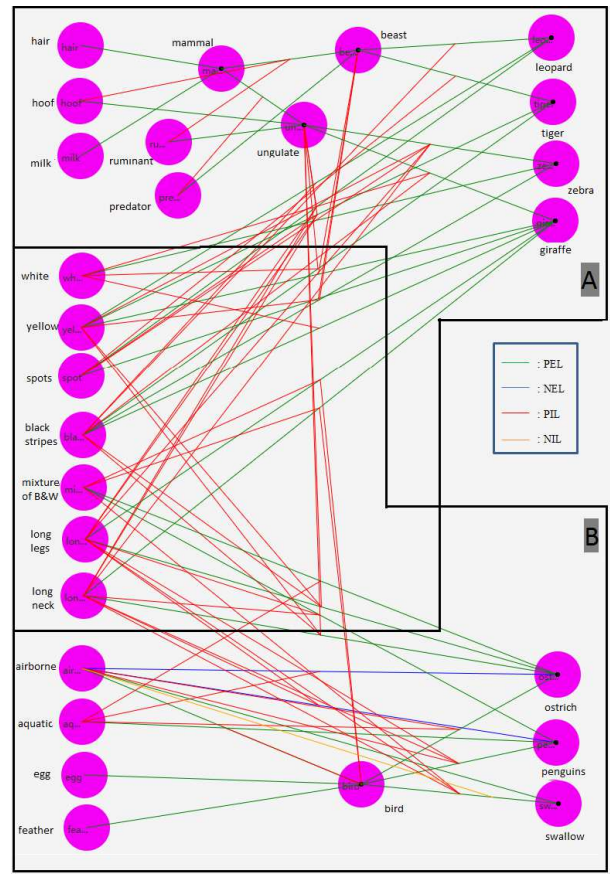}
\caption{Integrated Neural network structure to represent logical relations of animals including mammals and birds}
\label{animal_zone}
\end{center}
\end{figure}

As shown in Figure~\ref{animal}, Figure~\ref{mammal} is a neural network structure $N_{A}$ learning the logical relations in the domain of mammals, and Figure~\ref{bird} is a neural network structure $N_{B}$ learning the logical relations in the domain of birds. Figure(c) is an integrated neural network structure learning the logical relations in the domain of animals including mammals and birds. It can further integrate other neural network structures such as in the domain of fish, and become a neural network structure learning the logical relations in the domain of animals. From the integrated neural network structure $N_{I}$ shown in Figure~\ref{animal_zone}, the partial neural network structure regarding mammals is similar with that of $N_{A}$ specified for the domain of mammals, which shows that $N_{I}$ reuses the structure of $N_{A}$ as its own. So does it regarding the domain of mammals. The integrated neural network structure $N_{I}$ has more links added rather than simply putting $N_{A}$ and $N_{B}$ together so as to represent all the logical relations correctly after the integration. This shows the neural network structure has the ability of absorbing the existing knowledge and fusing them rather than restarting from zero so as to fulfil incremental learning in respect of representing logical relations.
\section{Conclusion}
Logical relations are a kind of basic and important relations in the world. In order to represent and store logical relations directly and increasingly into neural network structures, this paper proposed a novel neural network model named PLDNN specified for this purpose. In this model, neurons and links, the two components in neural network, are designed specified for representing logical relations. By combining these basic neural components, complex logical relations can be mapped directly into the neural network structure without extra hidden logical neurons and layers. Dynamical construction and adjustment methods of the neural networks are designed so as to make PLDNN can automatically and increasingly represent and store logical relations into the neural network along with the happening of events. In addition, the dynamical creation of neurons and links makes the network structure of PLDNN dynamical and unfixed so as to represent and store more information on logical relations by the connection structure(or called the shape of neural network), not just by the weight of links.

PLDNN can be used in automatically establishing of logical relations in the rule library in expert systems such as medical diagnostic systems, and it can continue updating its network structure along with the change of logical relations so as to represent and store these changes. In addition, integration can be made among multiple existing learned neural networks into one single big neural network so as to absorb and fuse the existing knowledge rather than restarting from zero in respect of representing logical relations. This paper provides a proto-type model of the neural network specified for representing and storing logical relations, as well as its automatic construction and adjustment methods, and the paper also makes some theoretical analysis of the relations between the neural network structure and the representation of logical relations. In the future, further researches on optimization will be carried out in the application of PLDNN in various domains.


%

\section*{Acknowledgment}
This work was supported by the National Natural Science Foundation of China (grant nos.61520106006 and 61133010).

\ifCLASSOPTIONcaptionsoff
  \newpage
\fi



%

\appendices
\section{}
\label{Supplements}
In the following, it is a simple rule library as example which contains 14 logical relations. The PLDNN in Figure~\ref{animal_zone} memorizes them and forms a knowledge graph through the interconnection structure of the neural network.
\begin{enumerate}
  \item If an animal has hair,  then it is mammal
  \item If an animal produces milk,  then it is mammal
  \item If a mammal is predator,  then it is beast
  \item If a mammal has hoof,  then it is ungulate
  \item If a mammal is ruminant, then it is ungulate
  \item If an animal has feather, produces egg,  then it is bird
  \item If an animal airborne, then it is bird
  \item If a beast is yellow and spots,  then it is leopard
  \item If a beast is yellow and black strips,  then it is tiger
  \item If an ungulate has long neck, long leg, yellow and spots, then it is giraffe
  \item If an ungulate is white and black strips,  then it is zebra
  \item If a bird cannot airborne, has long neck, long legs, and is mixture of black and white,  then it is ostrich
  \item If a bird cannot airborne, can aquatic, and is mixture of black and white, then it is penguin
  \item If a bird can airborne,  then it is swallow
\end{enumerate}

These relations are usually used as example in AI-related book and materials like Haykin's Neural Networks and Learning Machines.
\section{}
\label{algorithm}
The algorithms of the construction and adjustment of the neural network structure are as follows. In the algorithms, ActivatedSet is the set of the activated neurons representing things in $E_{pre}$. ActivatedELSet(o) is the set of ELs in the activated state whose pre-end neuron is o. RAILSet ($EL_{j}$) is the set of ILs in the activated state to inhibit $EL_{j}$. ExcitedSet is the set reasoned by $o_{i}$. PreActivatedELSet($o_{j}$) is the set of ELs whose post-end neuron is $o_{j}$. PActivatedSet is the set of the positively activated neurons. NActivatedSet is the set of the negatively activated neurons. PreNeuronSet is the set of the pre-end neurons of EL.
\begin{figure*}[!h]
\begin{center}
\includegraphics[width=0.9\textwidth]{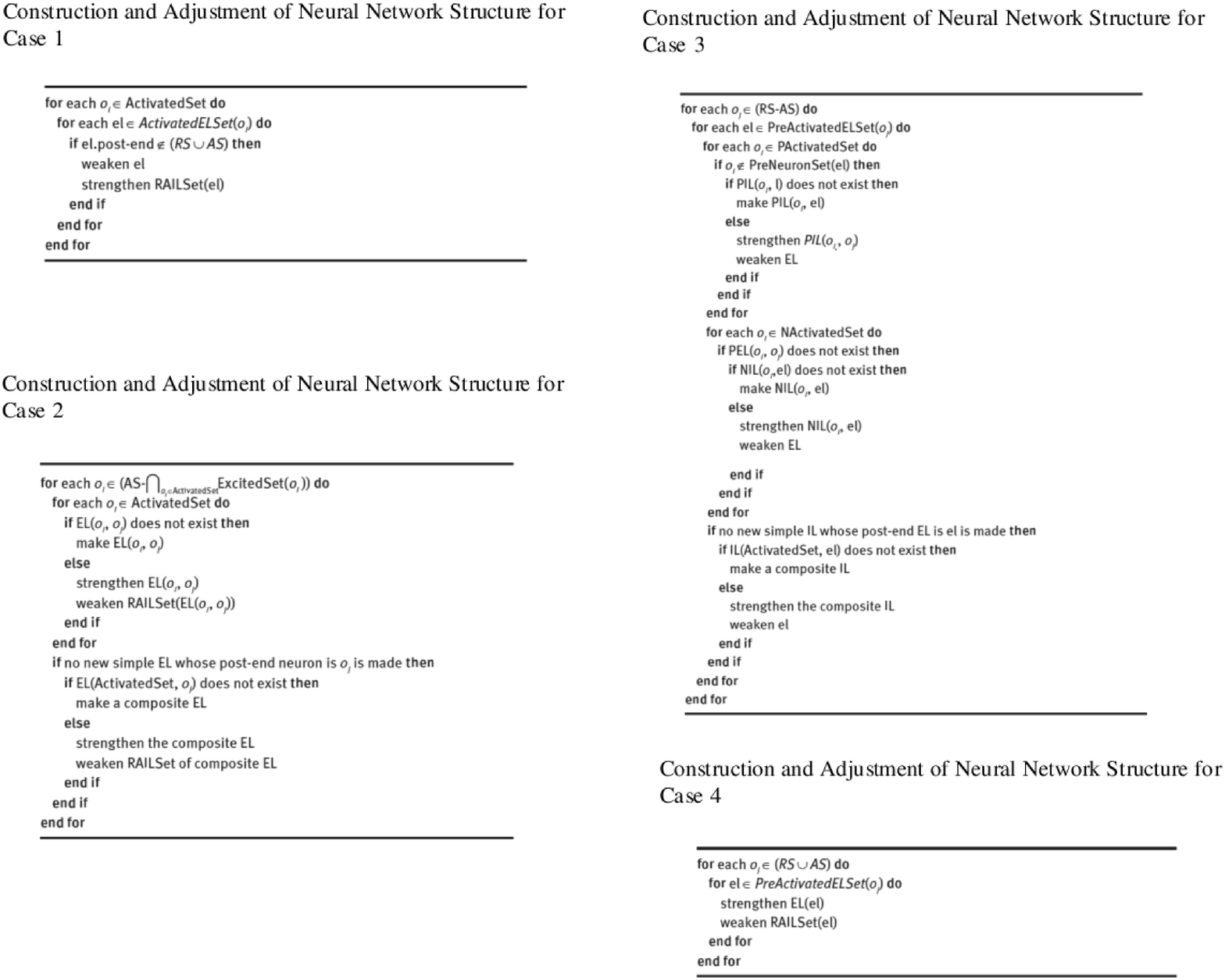}
\end{center}
\end{figure*}



\end{document}